\pdfoutput=1 
\documentclass[11pt,a4paper]{article}
\usepackage[hyperref]{emnlp2020}
\usepackage{times}
\usepackage{latexsym}
\usepackage[utf8]{inputenc} 
\usepackage[T1]{fontenc}
\usepackage{paralist}
\usepackage{tikz}
\usepackage{tikz-qtree}
\usepackage{tikz-dependency}
\usepackage{booktabs}
\usepackage{amsmath}
\usepackage{paralist}
\usepackage{lettrine}

\usepackage{url}
\newcommand{\rico}[1]{\todo[size=\tiny,color=blue!10]{\sffamily R: #1}}

\usepackage{microtype}

\aclfinalcopy 

\newcommand{\todon}[1]{\textcolor{red}{Nick: #1}}

\title{Domain, Translationese and Noise in Synthetic Data \\ for Neural Machine Translation}

\author{Nikolay Bogoychev$^1$ \quad Rico Sennrich$^{2,1}$ \bigskip\\
  $^1$School of Informatics, University of Edinburgh \\
  $^2$Institute of Computational Linguistics, University of Zurich \\
  \texttt{n.bogoych@ed.ac.uk, sennrich@cl.uzh.ch}
  }

\date{}

\begin{document}
\maketitle
\begin{abstract}
  Neural machine translation benefits from synthetic parallel training data. Source-side monolingual data can be (forward-)translated into the target language for self-training; target-side monolingual data can be back-translated. It has been widely reported that back-translation delivers superior results, but could this be due to artefacts in standard test sets? We perform a case study on a French-English news translation task and separate test sets based on their original languages. We show that forward translation can deliver superior BLEU in most cases on sentences that were originally in the source language, complementing previous studies which show large improvements with back-translation on sentences that were originally in the target language. However, humans mostly prefer the output of a backtranslated system, regardless of the input direction. We explore differences between forward and back-translation and find evidence for subtle domain effects, and show that the relative success of the two methods is affected by the quality of the systems used to produce synthetic data, with forward translation being more sensitive to this quality.
  \end{abstract}

\section{Introduction}

\begin{figure}[ht]
\centering
\scalebox{.7}{
  \begin{tikzpicture}
 \node[text width=5cm] at (-5, 2) {Original portion of test set\\ French news domain};
 \node[text width=5cm] at (1.5, 2) {Reverse portion of test set \\ English news domain};
\node[draw, text width=2.2cm] at (2,-4) (hu_fr) {Human \\ translationese French};
\node[draw, text width=2.2cm] at (2,0) (en_orig) {Original \\ English text};
\node[draw, text width=2.2cm] at (-2,-4) (mt_en) {Machine translationese English};
  \path [draw=black,->,thick] (en_orig) -- node [midway,above, sloped ] {Human} (hu_fr);
  \path [draw=black,->,thick] (hu_fr) -- node [midway,above, sloped ] {NMT} (mt_en);
  \path [draw=black,->,thick] (mt_en) -- node [midway,above, sloped ] {Compute BLEU} (en_orig);
\node[draw, text width=2cm] at (-6,0) (fr_orig) {Original \\ French text};
\node[draw, text width=2.2cm] at (-2,0) (hu_en) {Human \\ translationese English};
\node[draw, text width=2.2cm] at (-6,-4) (mt_en2) {Machine translationese English};

  \path [draw=black,->,thick] (fr_orig) -- node [midway,above, sloped ] {Human} (hu_en);
  \path [draw=black,->,thick] (fr_orig) -- node [midway,above, sloped ] {NMT} (mt_en2);
  \path [draw=black,->,thick] (mt_en2) -- node [midway,above, sloped ] {Compute BLEU} (hu_en);
  \end{tikzpicture}
  }
    \caption{Original and reverse portions of French$\to$English test set, differences in domain and differences in BLEU evaluation.}
    \label{translationese}
\end{figure}
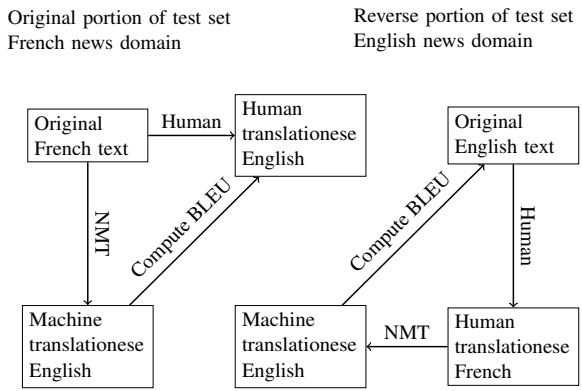

The quality of neural machine translation can be improved by leveraging additional monolingual resources in various different ways  \citep{sennrich-2016-BT, fwd-trans-orig, deep_shallow, google_pretrain, ape_directional}. Among these, back-translation is the most widely used technique in shared translation tasks \cite[p.\ 15]{wmt19}, and it has been reported that it outperforms self-training with forward translation \cite{burlot-fwd-bt}.
However, in the past year, attention was drawn to the fact that standard test sets are often shared between translation directions and thus contain both portions where the original text is on the source side (original), as well as portions where the original text is used as the reference translation, with the source text being a human translation (reverse) (See Figure \ref{translationese}). This use of ``original`` and ``reverse'' test sets heavily affects empirical results for back-translation.
When augmenting the model with back-translation, improvements in BLEU \citep{bleu} are a lot more evident if the sentence was translated in the reverse direction, that is to say with naturally produced reference and a human translation on the source side \citep{edunov2019evaluation}. \citet{ape_directional} explore automatic post editing (APE), which heavily relies on synthetic training data, and find that there is a loss in BLEU score on the original portion, despite humans perceiving improvement in the translation quality. \citet{translationese-effect-bleu-scores} show that the ranking of submissions to the news translation task changes when evaluating only the portion with original sources, or only that with translationese sources. Interestingly, systems that rely heavily on large-scale back-translation, such as that by \citet{edunov2019evaluation}, are more dominant on the reverse portion.

We focus on three factors that we hypothesise play a large role in explaining the observed differences in effectiveness between forward and back-translation, and between performance on the original and reverse portion of standard test sets: differences in language style between naturally produced text and translationese text, differences in the domains between source-side and target-side monolingual texts,\footnote{here, we use domain in a broad sense to refer to various textual attributes such as subject, genre, and topics.} and differences in how noise in the synthetic data, specifically translation errors due to the varying quality of MT systems producing it, will affect the final system, depending on whether it is on the source-side (back-translation) or target side (forward translation). We perform the following experiments to verify our claims:

\begin{itemize}
    \item We show that when the test sets are split according to original language, forward translation is generally better than backtranslation in terms of BLEU on the original portion, complementing the findings of \citep{edunov2019evaluation}, who find that back-translation is better at improving BLEU on the reverse portion.
    \item We perform human evaluation on a subset of our translation, comparing a baseline system, one augmented with back-translation and one augmented with forward translation. We see that despite the huge discrepancies in BLEU, humans measure adequacy to be pretty similar across all systems, especially on the original portion of the dataset. Humans, however tend to prefer the backtranslation's system fluency a lot more than that of the other two systems.
    \item We perform language model experiments where we contrast language style and language domain and evaluate on the test sets.
    \item We show that the language between original and translationese French is sufficiently different to be reliably detected by a neural network on a document level.
    \item We explore the effectiveness of forward and back-translation in a scenario where the quality of the synthetic data produced is poor, and find that forward translation is more sensitive to the quality of the initial translation system than back-translation.
    \end{itemize}

\section{Background}

Statistical machine translation relies on the noisy channel model, which makes large-scale language models, and hence extensive monolingual target-language data, very valuable \citep[e.g.][]{stupid_backoff}. In neural machine translation \citep{bahdanau_nmt, vaswani_transformer} however, it is not immediately clear how to make use of monolingual target-language resources. This led to the development of different methods such as language model fusion \citep{deep_shallow}, language model pretraining \citep{google_pretrain}, back-translation \citep{sennrich-2016-BT}, but also the exploration of methods to incorporate source-language data via forward translation \citep{fwd-trans-orig}. Out of these, back-translation is the most widely used \cite[see][]{wmt19}, and has been reported to work better than forward translation in particular \cite{burlot-fwd-bt}.


\subsection{Back-translation}

Given a translation task $L_1 \rightarrow L_2$, where large-scale monolingual $L_2$ data is available, back-translation refers to training a translation model $L_2 \rightarrow L_1$ and using it to translate the $L_2$ data into $L_1$, creating a synthetic parallel corpus that can be added to the true bilingual data for the purpose of training a $L_1 \rightarrow L_2$ model.

While this technique was first explored for statistical machine translation \citep{Bertoldi_Federico_2009,lambert-EtAl:2011:WMT,bojar-tamchyna-2011-improving}, it has a different effect on training, and was found to be much more effective, in neural machine translation, particularly in low resource scenarios \citep{sennrich-2016-BT,wmt16_edin}. However, it is not entirely clear what causes the large improvement in translation quality. Previous work has analysed increases in fluency when training on back-translated data \citep[e.g.][]{sennrich-2016-BT,edunov2019evaluation}, and domain adaptation effects \citep[e.g.][]{sennrich-2016-BT,chinea-rios-etal-2017-adapting}, which can be attributed to the target-side data, but the properties of synthetic source sentences have also been investigated. \citet{burlot-fwd-bt} have found that automatic translations tend to be more monotonic and simpler than natural parallel data, which could make learning easier, but these biases also make the training distribution less similar to natural input.
While there is some evidence that the quality of the back-translation system matters \citep{burlot-fwd-bt}, models are relatively robust to noise, and \citet{edunov-etal-2018-understanding} even find that they obtain better models when using sampling rather than standard beam search for back-translation, or explicitly add noise, even if this reduces the quality of back-translations. \citet{tagged_bt} argue that if the model is given means to distinguish real from synthetic parallel data, either via noise or more simply a special tag, it can avoid learning detrimental biases from synthetic training data.


\subsection{Forward translation}

Given a translation task $L_1 \rightarrow L_2$, where large-scale monolingual $L_1$ data is present, forward translation refers to training a translation model $L_1 \rightarrow L_2$ and using it to translate the $L_1$ data into $L_2$, creating a synthetic parallel corpus that can be added to the true bilingual data for the purpose of training an improved $L_1 \rightarrow L_2$ model.

Self-training with forward translation was also pioneered in statistical machine translation \citep{Ueffing2007}, but attracted new interest in neural machine translation, where improvements in BLEU were demonstrated \citep{fwd-trans-orig, fwd-trans}.
Compared to back-translation, biases and errors in synthetic data are intuitively more problematic in forward translation since they directly affect the gold labels. Also, there is no clear theoretical link between forward-translated synthetic training data and a model's fluency, but other effects, such as domain adaptation and improved learnability of translation from synthetic data remain plausible.\footnote{Also consider the effectiveness of sequence-level knowledge distillation \citep{kim-rush:2016:EMNLP2016}, which is similar to forward translation, except the source side of the parallel training data is re-translated, while we focus on integrating additional monolingual data.} 


\citet{burlot-fwd-bt} perform a systematic study which shows that forward translation leads to some improvements in translation quality, but not nearly as much as back-translation. In very recent work, \citet{fwd_bt_at_scale} show large-scale experiments where a combination of synthetic data produced by both forward and backward translation delivers superior results to just using one or the other. The amount of research on forward translation is however significantly smaller than that on back-translation.


\section{Domains and Translationese}

Based on these studies, we consider how the original and reverse portion of standard test sets differ, and how this can partially explain the observed differences between forward and back-translation.

\subsection{Domains}

It has previously been shown that back-translation can be used for domain adaptation \citep{sennrich-2016-BT,chinea-rios-etal-2017-adapting}, and the effectiveness of back-translation and forward translation heavily depends on the availability of relevant, in-domain monolingual data.
Even if we have both source-side and target-side data from the same general domain, we believe that there can be subtle differences between them. Even in restricted domain tasks, such WMT news translation \citep{wmt19}, newspaper articles in different languages talk about different topics.\footnote{Obviously, there will also be differences between newspapers in the same language, but we expect that a large-scale corpus from the same language will better match topics at test time than one from another language} For example, French news article cover subjects of local interest, such as the Quebec local elections. On the other hand, English language news in WMT test sets talk about mostly American or international topics. Therefore when performing back-translation, which is based on target-side data, this implicitly adapts systems to this target-side news domain, while forward translation would adapt systems to the source-side news domain.

\subsection{Translationese}

A second important distinction between the original and reverse portion of test sets comes from their creation, i.e.\ the process of translation. Human translations show systematic differences to natural text, and this dialect has been termed translationese.
 Translationese has been extensively studied in the context of natural language processing \citep{translationese-studies, translationese-studies-interpretese}. Translationese texts tend to have different word distribution than naturally produced text due to interference from the source language \citep{koppel-ordan-2011-translationese}, and other translation strategies such as simplification and explicitation. While translationese is hard to spot for humans, machine learning methods can reliably identify it \citep{spot_translationese,koppel-ordan-2011-translationese,unsupervised-translationese-id}.

Translationese and its effect have been studied in the context of statistical machine translation: \citet{Kurokawa2009AutomaticDO, statmt_directional} observe that systems reach higher BLEU on test sets if the direction of the test set is the same as the direction of the training set, \citet{translationese-study} show how systems can be tuned specifically to translationese and \citet{riley-etal-2020-translationese} even show how BLEU can be gamed by specifically producing translationese. Due to the directional nature of the WMT19 test sets \citep{wmt19}, research on translationese in the context of neural machine translation has been revitalized \citep{ape_directional, edunov2019evaluation, translationese-effect-bleu-scores,DBLP:journals/corr/abs-1906-09833,bizzoni-etal-2020-human}.

One of our goals in this paper is to isolate domain effects and translationese effects in the analysis of synthetic training corpora.

\section{Experimental setup}
\label{experiments}
We used the WMT 15 English-French news translation task dataset \citep{wmt15}, consisting of 35.8M parallel sentences. For back and forward translation we used 49.8M English monolingual sentences and 46.1M French monolingual sentences from the respective News Crawl corpora. For training the back-translation and forward translation systems we used a both a shallow RNN \citep{bahdanau_nmt}, equivalent to the one used by \citep{wmt16_edin}, as well as a transformer base system \citep{vaswani_transformer}. Our shallow RNN was about 1 BLEU better than the transformer on fr-en (used for forward translation), and about 3 BLEU worse on en-fr (used for backtranslation) than the transformer. For producing the synthetic data we used sampling from the softmax distribution \citep{edunov-etal-2018-understanding}. Byte pair encoding (BPE) \citep{subword_nmt} was used to produce a shared vocabulary of 88k tokens. 

For training the baseline model, as well as the ones augmented with synthetic data, we used the transformer base architecture. The models denoted as BT and FWD are trained by augmenting the parallel dataset with back-translation and forward translation respectively and a subscript $_{\text{transformer}}$ or $_{\text{rnn}}$ denotes what system was used to produce the synthetic data. All model hyperparameters are the same as those of the baseline. All training and decoding was done using the Marian machine translation toolkit \citep{mariannmt}. All models were trained with early stopping on a dev set (newstest2014) with patience 10.

\subsection{Directional test sets}

We used all available datasets from the news translation task and split them by direction, based on the source language, equivalent to the way done by \citet{sacreBLEU}, and we evaluated each dataset with all of our models.

\section{Translation experiments}

\begin{table}[ht]
\small
\begin{tabular}{lllllll}
\toprule
\textbf{System} & \multicolumn{1}{c}{\textbf{2008}} & \multicolumn{1}{c}{\textbf{2009}} & \textbf{2010} & \textbf{2011} & \textbf{2012} & \textbf{2013} \\
\midrule
\multicolumn{7}{l}{Original (French source)}                                                                                                              \\
Baseline                      & 29.4                     & 44.2                     & \textbf{32.9} & 32.3          & 37.3          & 47.4          \\
BT$_{\text{transformer}}$     & 28.0                     & 41.8                     & 30.0          & 30.3          & 34.0          & 45.8          \\
BT$_{\text{rnn}}$             & 29.3                     & 42.2                     & 31.7          & 31.5          & 34.6          & 46.9          \\
FWD$_{\text{transformer}}$    & 29.0                     & 43.8                     & 32.3          & 32.4          & 36.4          & 49.0           \\
FWD$_{\text{rnn}}$            & \textbf{30.9}            & \textbf{45.1}            & 32.0          & \textbf{33.1} & \textbf{38.3} & \textbf{48.3}        \\
\midrule
\multicolumn{6}{l}{Reverse (Translationese French source)}                                                                &               \\
Baseline                     & 29.1                 & 29.6                  & 37.3          & 45.3          & 34.5          & 35.4          \\
BT$_{\text{transformer}}$    & 31.6                 & 32.9                  & 42.6          & \textbf{50.8} & \textbf{39.3} & \textbf{39.5} \\
BT$_{\text{rnn}}$            & \textbf{32.1}        & \textbf{33.4}         & \textbf{43.3} & 50.5          & 39.0          & 38.4          \\
FWD$_{\text{transformer}}$   & 28.0                 & 28.7                  & 36.7          & 44.5          & 33.7          & 35.2         \\
FWD$_{\text{rnn}}$           & 27.5                 & 28.1                  & 36.0          & 43.0          & 33.0          & 33.9         \\
\midrule
\multicolumn{7}{l}{Full test set*}                                                                                                                      \\
Baseline                    & 29.2                  & 37.3                  & 35.2          & 38.8          & 35.9          & 41.6          \\
BT$_{\text{transformer}}$   & 30.0                  & 37.6                  & 36.1          & \textbf{40.7} & 36.8          & \textbf{42.9}   \\
BT$_{\text{rnn}}$           & \textbf{30.9}         & \textbf{38.1}         & \textbf{37.3} & 40.5          & \textbf{36.9} & 42.5   \\
FWD$_{\text{transformer}}$  & 28.5                  & 36.7                  & 34.4          & 38.5          & 35.0          & 42.2 \\
FWD$_{\text{rnn}}$          & 29.0                  & 37.1                  & 33.9          & 38.0          & 35.6          & 41.5 \\ 
\bottomrule
\end{tabular}
\caption{French$\to$English BLEU scores on newstest. \textbf{BT} and \textbf{FWD} denote baseline system augmented with transformer or RNN produced back-translation and forward translation respectively. \newline *For some test sets, some of the sentences are originally in neither French nor English, so we removed them from the test set.}
\label{full_trans_experiments}
\end{table}


We present our experimental results on Table \ref{full_trans_experiments}.
On the original portion, the systems augmented with FWD$_{\text{rnn}}$ translated data performs the best on most test sets. The back-translation system is worse than the baseline on all test sets. Furthermore it is interesting to observe that the system trained on transformer-produced synthetic data is worse than that trained on RNN-produced synthetic data.

We observe the opposite on the reverse portion: a back-translation system (either BT$_{\text{rnn}}$ or BT$_{\text{transformer}}$, with no clear winner between the two, despite a 3 BLEU difference between the quality of the baseline RNN and Transformer) is always the best, and the forward translation systems shows no improvement over the baseline.

On the full datasets, the overall trend is that forward translation does not improve the overall translation quality, which is not consistent with previous work \citep{burlot-fwd-bt}. We note that RNN produced synthetic data mostly outperforms their transformer counterparts. We note that overall, the backtranslation augmented system produces the best BLEU, which is consistent with \citet{burlot-fwd-bt}. It is tempting to conclude that forward translation works better for texts in the original translation direction, but we can't do that without conducting human evaluation, as BLEU is known to not correspond directly to translation quality, especially for high quality systems \citep{ma-EtAl:2019:WMT, freitag2020bleu, mathur-etal-2020-tangled}. It does seem that forward translation is more sensitive to the quality of the system used to produce the synthetic data. 


\section{Human Evaluation}
Table \ref{full_trans_experiments} shows big discrepancies in the BLEU scores based on the type of synthetic data and directionality of the datasets, but BLEU does not tell the full story. In order to get further insight on the effects of forward and backward translated data, we sampled uniformly 1008 sentences from all the newstest datasets, 504 in the forward direction and 504 in the reverse direction. We recruited 4 native English speakers to evaluate the translations of those sentences with three distinct systems (the baseline, BT$_{\text{rnn}}$, and FWD$_{\text{rnn}}$). We followed the evaluation scheme of \citet{meta-eval} where we request our annotators to rate translations in terms of fluency and adequacy on a scale from 1 to 5. 
Annotators are only shown the three translations for the fluency evaluation; for the adequacy evaluation, they are additionally provided with a reference translation.
Rating scales and instructions are shown in the Appendix.

Translations are blinded and given in random order to prevent biases.
 Each annotator was asked to annotate 377 sentences for fluency and adequacy each and the sets for fluency and adequacy are distinct. Among those, 50 sentences appear twice in order to measure intra-annotator agreement, and 100 sentences are common across all annotators in order to measure inter-annotator agreement. We report Kohen's Kappa scores \citep{kappa_scores} for annotator agreement on Table \ref{agreement}. 
We test statistical significance with three-way $p$-values computed using the ANOVA test \citep{anova_tst}. We also report results of the $t$-test, comparing the FWD and BT systems.

Our human evaluation results are presented in Table \ref{hueval}. In terms of adequacy on the original portion of the dataset, we see that all systems perform very similarly, with no significant differences between systems.
On the reverse portion of the dataset, backtranslation has a slight edge over the baseline, and a more notable edge against the forward translation system which is consistent with related work. In terms of fluency the results are more clear: The backtranslation system clearly produces more fluent output than its competitors, regardless of the translation direction. This finding is consistent with the findings of \citet{edunov-etal-2020-evaluation} who also show that humans have a preference for backtranslation augmented systems due to their more fluent output.

\begin{table}[ht]
\centering
\small
\begin{tabular}{@{\hskip3pt}l@{\hskip3pt}c@{\hskip3pt}c@{\hskip3pt}c@{\hskip3pt}c@{\hskip3pt}}
& \multicolumn{2}{c}{Adequacy} & \multicolumn{2}{c}{Fluency} \\
\cmidrule(l){2-3}
\cmidrule(l){4-5}
System   & Original & Reverse & Original & Reverse \\
\toprule
Baseline & \textbf{4.18}  & 4.05           & 3.87          & 3.96          \\
BT       & 4.15           & \textbf{4.14}  & \textbf{4.05} & \textbf{4.10} \\
FWD      & 4.10           & 3.98           & 3.60          & 3.7           \\
\midrule
$p$-value & 0.244          & $<$0.011       & $<$0.00001    & $<$0.00001 \\ 
\midrule
\multicolumn{5}{c}{$t$-test on FWD and BT} \\
\midrule
$p$-value  &  0.33        & 0.003           & $<$0.00001    & $<$0.00001 \\
\bottomrule
\end{tabular}
\caption{French$\to$English human evaluation. \iffalse Three way statistics test was computed using the ANOVA test \citep{anova_tst}. Standard $t$-test was used for the two way comparison between FWD and BT.\fi }
\label{hueval}
\end{table}

\begin{table}[ht]
\centering
\begin{tabular}{ccc}
Annotator   & Fluency & Adequacy \\
\toprule
0           & 0.28    & 0.50     \\
1           & 0.38    & 0.67     \\
2           & 0.47    & 0.61     \\
3           & 0.38    & 0.69     \\
\midrule
\{0 1 2 3\} & 0.16    & 0.26   \\ 
\bottomrule
\end{tabular}
\caption{K scores for intra- and inter- annotator agreement. For inter-annotator agreement we report average scores between the agreement of any two annotators. }
\label{agreement}
\end{table}

\section{Language model experiments}
BLEU scores are insufficient to draw conclusions about the nature of the improvements both data augmentation methods bring. We previously touched upon two hypotheses:

\begin{itemize}
    \item translationese effects: the references in the reverse portion are native-produced text, those in the original portion may contain translationese artifacts. Training on backtranslations may improve language modelling and favour the production of more native-like text, while training on forward translations may bias the MT system towards producing more translationese text.
    \item domain effects: there may be subtle domain differences in the synthetic data sets, mirroring differences between the two portions of the test set.
\end{itemize}

We designed a language modelling experiment in order to distinguish between the two explanations.
Specifically, we measure the similarity between training and test sets by training language models on our training data, and measuring perplexity to variants of the test sets.

We used a transformer language model architecture with 8 layers and 8 heads, similar to the transformer-base machine translation systems. We used the same preprocessing and BPE settings as our translation experiments.
We trained four language models using the data that we had prepared for forward and back-translation: two native English and French language models and two English and French translationese models (we denote the latter two EN$_\text{MT}$ and FR$_\text{MT}$, respectively).
The language models computed on the machine translated data exhibit specific features: They are trained on sampled data so we expect below average fluency, but good adaptation to the domain (source-side news or target-side news). Therefore we expect that the native French language model will perform better (i.e.\ have lower perplexity) on native French text compared to a translationese French language model, as the style and the domain of the native text match with those of the native language model. We expect that we will observe the same effect when evaluating native vs translationese English language model on native English text.
When considering translated test sets, we will expect them to be closer to the translationese language models -- this is both compatible with the interpretation that the two types of texts are similar because they are both translationese, as well as the interpretation that they are similar because they are from the same source-language domain.

But what if we have native English data that has been human translated into French and then automatically translated into English? In this case it will share the domain with native English, but after the intermediate human translation, we expect the style to be closer to the language model trained on the translationese text. This variant of the test set gives us the most direct answer as to what extent translationese or domain effects affect the similarity between training and test data.

\begin{table}[ht]
\centering
\small
\begin{tabular}{lcc}
\toprule
 & \multicolumn{2}{c}{\bf LM}\\
\cmidrule(l){2-3}
\textbf{test set} & \bf FR$_\text{native}$   & \bf FR$_\text{MT}$\\
\midrule
native FR                & \textbf{\phantom{0}99.22}  & 118.83           \\
HT$_{\text{EN}\to\text{FR}}$           & \textbf{113.98} & 117.97          \\
\bottomrule
\end{tabular}
\caption{Language model perplexity on the French side of the combined directional datasets, normalised by number of sentences. We distinguish whether test sets are native French or human translation (HT).}
\label{frlm}
\end{table}

\begin{table}[ht]
\centering
\small
\begin{tabular}{lcc}
\toprule
& \multicolumn{2}{c}{\bf LM}\\
\cmidrule(l){2-3}
\textbf{test set} & \bf EN$_\text{native}$   & \bf EN$_\text{MT}$ \\
\midrule
native EN               & \textbf{101.90} & 118.71           \\
HT$_{\text{EN}\to\text{FR}}$, MT$_{\text{FR}\to\text{EN}}$      & \textbf{\phantom{0}98.01}  & \phantom{0}99.71\\  
\midrule
MT$_{\text{FR}\to\text{EN}}$           & 102.28          & \textbf{\phantom{0}94.43}   \\
HT$_{\text{FR}\to\text{EN}}$           & 113.99          & \textbf{111.90}  \\
\bottomrule
\end{tabular}
\caption{Language model perplexity on the English side of the combined directional datasets, normalised by number of sentences. We distinguish whether test sets are native English, human translation (HT), machine translation (MT), or roundtrip translation with multiple translation steps.}
\label{enlm}
\end{table}

Table \ref{frlm} shows the language model performance of the native French language model and the translationese French language model. We observe that unsurprisingly, the language model trained on original French data shows lower perplexity on the original French data than the one trained on MT translated French. Somewhat surprisingly the trend is maintained in the translationese French dataset, even if the two perplexity scores are closer to each other. This is unlike the results on the English language models on Table \ref{enlm}, where the language model that performs better is always the one trained on the same original language as the original language of the dataset.

Of most interest are the result for HT$_{\text{EN}\to\text{FR}}$, MT$_{\text{FR}\to\text{EN}}$, i.e.\ the roundtrip translation of native English text.
Based on our hypothesis that source-language and target-language domains are slightly different, we expect the EN$_\text{native}$ LM to perform better than EN$_\text{MT}$.
Based on the more established explanation that the main distinguishing feature of translated text are translationese artefacts, we would expect EN$_\text{MT}$ to perform better than EN$_\text{native}$.
In fact, perplexities are very close to each other, suggesting that domain effects and translationese effects both come into play, and roughly balance each other out.

\section{Domain identification experiments}
Inspired by the work of \citet{tagged_bt, marie-etal-2020-tagged}, who tag back-translated data on the source side to distinguish it from parallel data, we explore if translation models can learn whether training instances come from the source-language or target-language ``domain''.
To this end, we train a French$\to$English translation model only using synthetic training data (both forward translations and back-translations), and we add a tag at the beginning of the target sentence indicating the original language.
The resulting model correctly identifies the original language in 83\% of training set sentences.
When evaluating it on test sets, the model has a marked preference to identify the original language as French.
On the originally French portion, the model found 89.4\% of the sentences be native French, whereas on the human translated French portion, the model predicts 51\% of the sentences to be native French.

\citet{tagged_bt} motivate source-side tags as a way to help the system distinguish back-translations from  parallel text in lieu of noise.
While we did not test the effectiveness of source-side tags, our experiment shows that even a model without them can predict the provenance of source sentences relatively well.
The fact that prediction accuracy remains far above chance level (69\%) on human translations shows that the high classification accuracy cannot be simply explained by the model learning to identify MT noise; the signal the model uses to correctly classify the test sets are either domain effects, or translationese effects shared between human translation and MT.

\section{Other Language Pairs}
To see if our findings generalise to other language pairs, we trained Estonian$\to$English and Finnish$\to$English translation models, following the procedure described in Section~\ref{experiments}. In order to better control for domain and style, we only use the parallel news crawl data from the WMT18 \citep{wmt18} translation task, which resulted in 3.1M sentence pairs for Finnish--English and 0.9M sentence pairs for Estonian--English. 

For data augmentation, we use all the available news-crawl on the Estonian/Finnish side for forward translation and the equivalent amount of English newscrawl for back-translation. This resulted in 14.5M monolingual sentences for Finnish-English back/forward translation and 2.9M sentences for Estonian-English back/forward translation. We again produced an RNN and a transformer variant of the synthetic data.
\begin{table}[ht]
\centering
\small
\begin{tabular}{lll}
\textbf{System} & \textbf{2018dev} & \textbf{2018test} \\
\toprule
\multicolumn{3}{c}{Original (Estonian source)}    \\
\midrule
Baseline        & 18.0          & 19.4   \\
BT$_{\text{rnn}}$              & 17.1          & 17.9    \\
BT$_{\text{transformer}}$       & \textbf{20.8}          & \textbf{21.5}    \\
FWD$_{\text{rnn}}$             & 16.2          & 17.4  \\
FWD$_{\text{transformer}}$              & 19.6          & 20.8  \\
\midrule
\multicolumn{3}{c}{Reverse (Translationese Estonian source)}    \\
\midrule
Baseline        & 20.2            & 20.6  \\
BT$_{\text{rnn}}$              & 23.2            & 22.8  \\
BT$_{\text{transformer}}$   & \textbf{29.4}            & \textbf{28.0}  \\
FWD$_{\text{rnn}}$             & 17.9            & 18.3     \\
FWD$_{\text{transformer}}$              & 20.5            & 20.8     \\
\midrule
\multicolumn{3}{c}{Full test set}   \\
\midrule
Baseline        & 19.1             & 20.0    \\
BT$_{\text{rnn}}$              & 20.1             & 20.5   \\
BT$_{\text{transformer}}$      & \textbf{25.1}             & \textbf{24.9}   \\
FWD$_{\text{rnn}}$             & 17.1             & 17.8   \\
FWD$_{\text{transformer}}$              & 20.6             & 20.8   \\
\bottomrule
\end{tabular}
\caption{BLEU scores on Estonian$\to$English. The RNN and Transformer subscripts refer to the system used for producing backtranslation.}
\label{estonian}
\end{table}

\begin{table}[ht]
\centering
\small
\begin{tabular}{lllll}
\textbf{System} & \textbf{2015} & \textbf{2016} & \textbf{2017} & \textbf{2018} \\
\toprule
\multicolumn{5}{c}{Original (Finnish source)}    \\
\midrule
Baseline        & \textbf{22.6}          & 24.2    & \textbf{24.2}   & 19.4  \\
BT$_{\text{rnn}}$              & 20.9          & 21.7    & 21.7   & 19.4 \\
BT$_{\text{transformer}}$              & 21.5          & 22.9    & 22.2   & \textbf{20.0} \\
FWD$_{\text{rnn}}$             & 19.6          & 22.2    & 22.4   & 13.9  \\
FWD$_{\text{transformer}}$             & 22.4          & \textbf{24.3}    & \textbf{24.2}   & 15.0  \\
\midrule
\multicolumn{5}{c}{Reverse (Translationese Finnish source)}    \\
\midrule
Baseline        & 18.9      & 22.7    & 26.1   & 22.1 \\
BT$_{\text{rnn}}$              & 22.6      & 28.8    & 31.5   & 20.3 \\
BT$_{\text{transformer}}$              & \textbf{24.1}      & \textbf{30.4}    & \textbf{33.3}   & 21.4 \\
FWD$_{\text{rnn}}$             & 16.8      & 20.4    & 23.4   & 20.1    \\
FWD$_{\text{transformer}}$             & 18.3      & 22.5    & 26.0   & \textbf{22.2}    \\
\midrule
\multicolumn{5}{c}{Full test set}   \\
\midrule
Baseline        & 20.6    & 23.4    & 25.2   & 18.3 \\
BT$_{\text{rnn}}$              & 21.9    & 25.7    & 27.2   & 19.8 \\
BT$_{\text{transformer}}$              & \textbf{23.0}    & \textbf{27.1}    & \textbf{28.4}   & \textbf{20.6} \\
FWD$_{\text{rnn}}$             & 18.1    & 21.2    & 22.9   & 16.5  \\
FWD$_{\text{transformer}}$             & 20.2    & 23.3    & 25.2   & 18.1 \\
\bottomrule
\end{tabular}
\caption{BLEU scores on Finnish$\to$English. The RNN and Transformer subscripts refer to the system used for producing backtranslation.}
\label{finnish}
\end{table}

We present our results in Tables \ref{estonian} and \ref{finnish}. In the case of Estonian (Table \ref{estonian}), we have a scenario which produced particularly poor synthetic data: The RNN English--Estonian, reaches just 12 BLEU on the dev set, while the transformer---18 BLEU. On Estonian-English the RNN reaches 15 BLEU, while the transformer---17. We see that when BLEU is low, the quality of the synthetic data is much more important: The systems augmented with transformer back-translation gained 4.7 BLEU points on average against the RNN back-translation. Relatively, the forward translation system has improved significantly more: Just 2 BLEU points of difference between the RNN and transformer models used to create the synthetic data resulted in 3.2 points increase in BLEU. This suggests that data augmentation via forward translation is substantially more sensitive to the translation quality of the initial translation system than back-translation.

Our observations are confirmed in the slightly higher-resource experiment on Finnish$\to$English (Table~\ref{finnish}). The quality of the translation model used for back-translation was improved by 9 BLEU (from 17 to 26) when using a transformer instead of an RNN, but on the final system, this yielded just 1.1 BLEU increase on average. In contrast, the quality of the translation system used for forward translation was improved from 17 to 23 BLEU, improving the final system by 2 BLEU on average.


\section{Conclusions}
In this paper we reviewed the effect of directionality on machine translation results, focusing both on the direction of data augmentation (forward and back-translation), and the original language of test sets, focusing on French$\to$English as a case study, with additional experiments on Estonian$\to$English and Finnish$\to$English. We confirm that the original language of parallel test sets affects BLEU scores, particularly when data augmentation approaches are compared. We find that back-translation is more effective than forward translation in the artifical setting where the input to the translation system is itself a human translation, and the original text is used as reference. In the natural setting where the input is native text, and the reference a human translation, forward translation can perform better in terms of BLEU, although it still trails behind back-translation if the forward translations in the synthetic data sets are very poor, indicating that forward translation is more sensitive to the quality of the system that produced it compared to backtranslation.

However, manual evaluation shows that better BLEU scores do not necessarily correspond to better translation quality according to human judgements. Despite wildly differing BLEU results depending on the original language of the test sets, humans evaluators prefer our backtranslated systems over our other systems in terms of fluency.
Despite achieving higher BLEU on the original portion of the test set, our forward-translation system was rated worst in the human evaluation.


To better understand the differences between forward and back-translation, we consider both translationese effects and subtle domain differences between source-language and target-language monolingual data. Language model experiments indicate that both of these play a role, and partially explain why back-translation is so suitable for reverse test sets. Experiments with translation systems trained on only synthetic data (forward and back-translation) also show that the provenance of test set sentences is predictable with 69\% accuracy.

Our findings are agree with concurrent and independent work by \citet{sourcetargetFB}, who perform low-resource translation experiments with back-translation and self-learning, an iterative form of forward translation. They also find that the original language of parallel test data determines whether back-translation or forward translation is a more effective strategy for data augmentation.


Based on our findings, we can make several recommendations for the use of forward translation and back-translation to augment neural machine translation.
Firstly, while BLEU is very sensitive to the choice of data augmentation, with up to 6 BLEU difference between the two choices in our French$\to$English experiments, depending on the make-up of the test set, human annotators were less sensitive to test set directionality.
Human annotators favoured backtranslation over forward translation, mostly in terms of fluency, while adequacy was largely the same across all of them, especially on the original translation direction. Our results should serve as a warning to not over-rely on automatic evaluation when data augmentation is involved: The results of the WMT19 news translation task \citep{wmt19}, also show negative correlation between BLEU and human evaluation.


Secondly, we observe subtle domain differences between corpora in different languages, even if they cover the same general domain (news) and were collected with the same methods. Following the general heuristic to use training data that matches the test domain as closely as possible, this may be an argument for using forward translations in settings where (only) in-domain source data is available, but further study of this setting is necessary. 
Of course, the use of forward and back-translation is not mutually exclusive, and in settings with access to suitable monolingual corpora in both the source and target language, combining the two is another viable strategy \citep{fwd_bt_at_scale}.

\bibliography{emnlp2020}
\bibliographystyle{acl_natbib}
\clearpage
\onecolumn
\appendix

\section{Human Evaluation Protocol}

The annotators were given the following instructions:

Fluency is simply how natural a sentence sounds. You will have to rate the sentences produced by three different systems in terms of how good the English is. Use the following scale:
\begin{compactitem}
    \item[\textbf{1}] Incomprehensible
    \item[\textbf{2}] Disfluent English
    \item[\textbf{3}] Non-native English
    \item[\textbf{4}] Good English
    \item[\textbf{5}] Flawless English
\end{compactitem}

Adequacy on the other hand, tries to judge how good the meaning is conveyed, compared to a reference translation. For this task, you will be given a reference translation, and the translations produced by three different systems. You are required to rate each of them using the following scale:

\begin{compactitem}
    \item[\textbf{1}] None
    \item[\textbf{2}] Little
    \item[\textbf{3}] Much
    \item[\textbf{4}] Most
    \item[\textbf{5}] All
\end{compactitem}

Please pay special attention to sentences that are "almost" correct but a crucial word is missed or one with reverse meaning is used (eg "convicted" vs "acquitted").

Note that it is possible to have all systems produce equally good (or equally bad), yet different results. For this task, you should ignore how fluent (or disfluent) the English is and focus just on the meaning. You should not penalise a system for having a bad language, as long as the meaning is conveyed.

\section{Manual analysis}
In this section we present manual analysis of sentences produced by all of our systems on Original French source (Figure \ref{orig_french_examples}) and on Translationese English source and (Figure \ref{orig_english_examples}). We noticed that in the case of original French source, sometimes the backtranslation system gets unfavourably penalised for producing a correct translation, that is more fluent than the reference (Figure \ref{orig_french_examples}, example 1). When the backtranslation is presented with an input that is confusing to it, it tends to just copy, instead of producing a translations (Figure \ref{orig_french_examples}, examples 2 and 3), whereas the forward translation system never suffers from this issue on original French source. 

On Translationese French source, we can see that the forward translation struggles with undertranslation on some out of domain sequences (Figure \ref{orig_english_examples}, example 7). It is interesting to note that, rare named entities such as "Xindu" on example 5, get translated as "Xinhua" by the backtranslation system, because this token is much more common there than in the training set (and completely absent in the forward translated data). Strangely enough, the backtranslation system tends to hallucinate "he said" when it translates quotes (Figure \ref{orig_english_examples}, example 6).

\begin{figure*}[hb]
\centering
\small
\begin{tabular}{lp{14cm}}
\toprule
source & Espérons que cela ne sera pas le cas.\\
reference & Let us hope that this is not the case.\\
\midrule
baseline & Let us hope that will not be the case.\\
FWD & Let us hope that this will not be the case.\\
BT & Hopefully that won't happen.\\
\bottomrule
\bottomrule
source & Mais ça n'est pas l'entrée en lice de Ségolène Royal qui pousse Nicolas Sarkozy à se dévoiler.\\
reference & But it is not the entry into the lists of Segolene Royal, which pushes Nicolas Sarkozy to unmask himself.\\
\midrule
baseline & But it was not the entry of Ségolène Royal that led Nicolas Sarkozy to discover himself.\\
FWD & But it is not the entry into the contest of Ségolène Royal, which pushes Nicolas Sarkozy to unveil himself.\\
BT & Mais ça n'est pas l'entrée en lice de Ségolène Royal qui pousse Nicolas Sarkozy à se prononcer.\\
\bottomrule
\bottomrule
source & L'Eure, l'Eure-et-Loir, les Deux-Sèvres et la Vienne, se sont ajoutés aux 20 départements déjà en vigilance orange, 18 depuis mercredi après-midi et deux, la Dordogne et la Gironde, plus tard dans la nuit.\\
reference & Eure, Eure-et-Loir, Deux-Sèvres ad Vienne have joined the 20 départements already on orange alert, 18 since Wednesday afternoon and two, Dordogne and Gironde, later during the night.\\
\midrule
baseline & the Eure , the Eure and Loir , the Two Shares and Vienna have joined the 20 departments already under orange vigilance , 18 since Wednesday afternoon and two , the Dordogne and the Gironde , later in the night .\\
FWD & The Eure, Eure-et-Loir, the Two-Shares and Vienna have been added to the 20 departments already in orange vigilance, 18 since Wednesday afternoon and two, Dordogne and Gironde later in the night.\\
BT & L'Eure, l'Eure-et-Loir, les Deux-Sèvre et la Vienne, se sont ajouté aux 20 départements déjà en vigilance orange, 18 depuis mercredi après-midi et deux, la Dordogne et la Gironde, plus tard dans la nuit.\\
\bottomrule
\bottomrule
source & Le but des futurs acquéreurs est de rouvrir les trois salles le 27 février prochain en projetant le très beau film de Xavier Beauvois Des hommes et des dieux.\\
reference & The aim of future purchasers is that of reopening the three halls on 27 February, showing the very beautiful film by Xavier Beauvois entitled "Men and Gods."\\
\midrule
baseline & The goal of future buyers is to reopen the three rooms on February 27 by screening Xavier Beauvus's very beautiful film Men and gods.\\
FWD & The purpose of future buyers is to reopen the three rooms on February 27 by projecting Xavier Beauvoir's beautiful film Les hommes et des gods.\\
BT & The goal of future buyers is to re-open the three rooms on February 27 next year with Xavier Beauvoir's beautiful film Men and Gods.\\
\bottomrule
\bottomrule
source & "Ce qui est intéressant, c'est que j'ai progressé à chaque compétition."\\
reference & "What is interesting is that I did better at each competition."\\
\midrule
baseline & "What's interesting is that I made progress in every competition."\\
FWD & "What is interesting is that I have made progress at each competition."\\
BT & "What's interesting is that I've progressed to every competition," he said.\\
\bottomrule
\bottomrule
source & Le nombre de chômeurs s'établit à 2,631 millions en métropole.\\
reference & The number of unemployed is 2.631 million in metropolitan France.\\
\midrule
baseline & The number of unemployed in metropolitan areas is 2,631 million.\\
FWD & The number of unemployed is 2,631 million in metropolitan France.\\
BT & The number of unemployed is 2.631 million in metropolitan areas.\\
\bottomrule

\end{tabular}
\caption{Original French source: BT produces natural paraphrase, but is penalised by BLEU.}
\label{orig_french_examples}
\end{figure*}


\begin{figure*}[hb]
\centering
\small
\begin{tabular}{lp{14cm}}
\toprule
source & Lundi soir, Peek et Rickard ont vérifié dans le motel.\\
reference & On Monday night, Peek and Rickard checked into the motel.\\
\midrule
baseline & He checked in the motel on Monday evening.\\
FWD & On Monday evening, Peek and Rickard checked in the motel.\\
BT & On Monday night, Peek and Rickard checked into the motel.\\
\bottomrule
\bottomrule
source & Il a même ordonné à M. Su, un pragmatique, de ne pas briguer la présidence.\\
reference & He even ordered Mr Su, a pragmatist, not to run for the presidency.\\
\midrule
baseline & He even ordered Mr President, a pragmatic one, not to run for the presidency.\\
FWD & He has even ordered a pragmatic President to not run the presidency.\\
BT & He even ordered Mr Su, a pragmatic, not to run for the presidency.\\
\bottomrule
\bottomrule
source & C'est bien."\\
reference & That is good."\\
\midrule
baseline & It is good."\\
FWD & That is good."\\
BT & It's good.""\\
\bottomrule
\bottomrule
source & Dans un jugement de 1966 qui a fait date, la Cour suprême des États-Unis a annulé la condamnation d'un ostéopathe de Cleveland, le Dr Sam Sheppard, pour l'assassinat de sa femme, soutenant dans sa décision qu'il y avait une "atmosphère de carnaval au procès " à cause des médias.\\
reference & In a landmark 1966 ruling, the United States Supreme Court overturned the conviction of a Cleveland osteopath, Dr. Sam Sheppard, for the murder of his wife, saying in its decision, Sheppard v. Maxwell, that there was a "carnival atmosphere at trial" because of the news media.\\
\midrule
baseline & In a landmark 1966 judgement, the Supreme Court of the United States set aside the conviction of an osteopcommitted by him for the murder of his wife. In its decision, the Supreme Court of the United States ruled that there was a "carnaval atmosphere at trial" because of the media.\\
FWD & In a 1966 judgement, the Supreme Court of the United States set aside the conviction of a Cleveland osteopath, Dr. Sam\\
BT & In a 1966 ruling, the U.S. Supreme Court overturned the conviction of a Cleveland osteopath, Dr. Sam, for the murder of his wife, arguing in his decision that there was a "carnival atmosphere at trial" because of the media.\\
\bottomrule
\bottomrule
source & Xindu a jusqu'ici échappé à une telle violence.\\
reference & Xindu has so far escaped such violence.\\
\midrule
baseline & So far, such violence has been avoided.\\
FWD & Xunfair has so far escaped such violence.\\
BT & Xinhua has so far escaped such violence.\\
\bottomrule
\bottomrule
source & "Luzhkov est au sommet."\\
reference & "Luzhkov is at the top."\\
\midrule
baseline & "Luzhkov is at the top."\\
FWD & "Luzhkov is at the top."\\
BT & "Luzhkov is at the top, he said." \\
\bottomrule
\bottomrule
source & Les quatre chiens d'exposition - deux akitas et deux corgis pembroke gallois - ont disparu mardi lorsque quelqu'un a volé la camionnette Chevrolet Express dans laquelle ils avaient passé la nuit à l'extérieur d'un motel 6 de Bellflower, ont annoncé les autorités.\\
reference & The four show dogs - two Akitas and two Pembroke Welsh corgis - vanished Tuesday when someone stole the 2006 Chevrolet Express van they had spent the night in outside a Bellflower Motel 6, authorities said.\\
\midrule
baseline & The four display dogs - two ak\\
FWD & The four exposure dogs - two akitas and two corgis\\
BT & The four showbiz dogs - two akitas and two Welsh Corgis - disappeared Tuesday when someone stole the Chevrolet Express van in which they had spent the night outside a Bellflower motel 6, authorities said.\\
\bottomrule
\end{tabular}
\caption{Translationese French source: BT produces natural paraphrase, but is penalised by BLEU.}
\label{orig_english_examples}
\end{figure*}

\end{document}